\documentclass[letterpaper, 10 pt, conference]{ieeeconf}  

\usepackage{cite}
\usepackage{graphicx}
\usepackage{booktabs}
\usepackage{hyperref}
\usepackage{amsmath} 
\usepackage{amssymb}

\IEEEoverridecommandlockouts                              

\title{\LARGE \bf
DexKnot: Generalizable Visuomotor Policy Learning \\ for Dexterous Bag-Knotting Manipulation
}

\author{
\authorblockN{Jiayuan Zhang$^{*}$, Ruihai Wu$^{*}$, Haojun Chen, Yuran Wang}
\authorblockN{Yifan Zhong, Ceyao Zhang, Yaodong Yang$^{\dagger}$, Yuanpei Chen$^{\dagger}$}
\authorblockA{Peking University\\
$^*$Equal contribution \quad $^\dagger$Corresponding author\\
\texttt{yaodong.yang@pku.edu.cn} \quad \texttt{yuanpei.chen312@gmail.com}}
}

\begin{document}

\maketitle
\thispagestyle{empty}
\pagestyle{empty}

\begin{abstract}

Knotting plastic bags is a common task in daily life, yet it is challenging for robots due to the bags' infinite degrees of freedom and complex physical dynamics. Existing methods often struggle in generalization to unseen bag instances or deformations. To address this, we present DexKnot, a framework that combines keypoint affordance with diffusion policy to learn a generalizable bag-knotting policy. 
Our approach learns a shape-agnostic representation of bags from keypoint correspondence data collected through real-world manual deformation.
For an unseen bag configuration, the keypoints can be identified by matching the representation to a reference. These keypoints are then provided to a diffusion transformer, which generates robot action based on a small number of human demonstrations. DexKnot enables effective policy generalization by reducing the dimensionality of observation space into a sparse set of keypoints. Experiments show that DexKnot achieves reliable and consistent knotting performance across a variety of previously unseen instances and deformations.

\end{abstract}


\begin{figure}[!t]
\centerline{\includegraphics[scale=0.58]{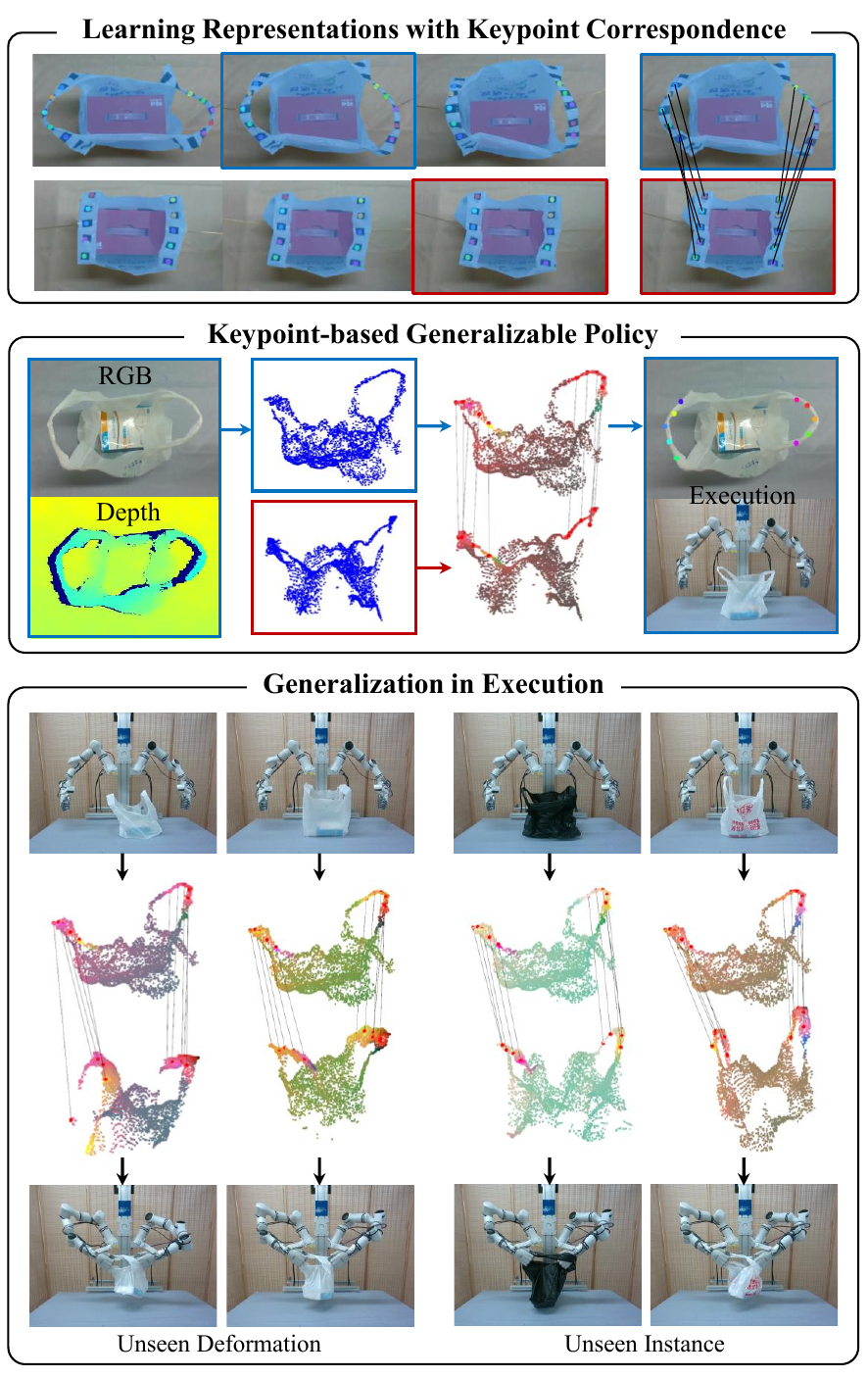}}
\caption{\textbf{Overview of DexKnot.} \textbf{Top row}: Our framework collects keypoint correspondence data through real-world manual deformation, which are used to learn shape-agnostic representations. \textbf{Middle row}: For a novel bag configuration, the keypoints are identified via correspondence matching, which guides the policy to execute the knotting task. \textbf{Bottom row}: Our framework generalizes effectively to unseen deformations and bag instances.}
\label{fig1}
\end{figure}

\section{INTRODUCTION}

Knotting plastic bags is a common and useful task in daily life, yet it is not easy for robots to handle such highly deformable objects \cite{yin2021modeling, chen2022autobag, chen2023bagging, bahety2023bag, gao2023iterative}.
In robot manipulation, while significant progress has been made in handling rigid and articulated objects \cite{geng2023partmanip}, operating deformable objects remains a formidable challenge for two primary reasons. 
First, their infinite degrees of freedom (DoF) lead to a very high-dimensional observation space, causing difficulties for a policy to learn and generalize. 
Second, deformable objects have complex and highly variable mechanical properties and physical dynamics, which are difficult to learn for neural surrogate models or to simulate in commonly used physical simulators.

Extensive research has explored the manipulation of deformable objects, including 1-dimensional (1D) lines like ropes \cite{nair2017combining, sundaresan2020learning, suzuki2021air}, 2-dimensional (2D) surfaces like clothes \cite{chen2022efficiently, weng2022fabricflownet, wu2024unigarmentmanip, wang2025dexgarmentlab}, and 3-dimensional (3D) volumetric bodies like plasticine \cite{shi2024robocraft}. 
In comparison, plastic bags \cite{gao2023iterative} present even greater challenges. Geometrically, bags exhibit hollow 3D structures \cite{seita2021learning} with openings and often contain internal items, requiring more precise and fine-grained manipulation. Dynamically, their softer, highly compliant materials lead to less structural stability. 
For instance, achieving even simple goal configurations, such as an upright pose, can be difficult, as bags tend to gradually collapse under their own weight. 
Existing studies on bag manipulation predominantly operate cloth bags without handles focusing on simple tasks, such as bag opening and object insertion. In these settings, bags can be treated as a folded cloth, simplifying both robot manipulation and physical simulation. 
However, the problem of knotting plastic bags, particularly generalizing across diverse bag instances and initial deformations, remains largely unexplored.

Despite their diverse deformations and sizes, most plastic bags share consistent topological structures (i.e., handles and openings) that enable us to learn invariant representations. This structural consistency also enables us to capture key features that are essential for manipulation while ignoring irrelevant details, motivating our design of a low-dimensional representation scheme. Additionally, the significant sim-to-real gap necessitates a real-world data collection pipeline rather than relying on physical simulation.

With these insights, we present DexKnot, a real-world policy learning framework for generalizable bag knotting, which leverages shape-agnostic contrastive representation learning, keypoints identification through correspondence matching, and keypoint-based generalizable diffusion policy (\autoref{fig1}).
We choose keypoints as representation because they reduce the dimensionality of the observation space, thus enhancing generalization especially when there are only a few demonstrations. 
The pipeline of our approach is as follows.
First, we perform real-world manual deformation to collect keypoint correspondence data across various bag instances and deformations. 
Next, we train a PointNet++ \cite{qi2017pointnet++} encoder to learn shape-agnostic representation of the bags' point clouds, allowing us to identify keypoints for an unseen bag configuration. The keypoints are taken as input by a diffusion transformer (DiT) \cite{peebles2023scalable}, which generates robot joint angle sequences trained with a few human demonstrations.

We evaluate DexKnot's performance and generalization capacity through systematic experiments. The results show that DexKnot has high success rates on both seen and unseen deformations for various seen and unseen bag instances. Compared to 3D Diffusion Policy (DP3) \cite{Ze2024DP3}, the state-of-the-art imitation learning framework, our approach demonstrates better generalization capacity on out-of-distribution deformations, such as twisted and inclined handle states.  

The main contribution of this work is the development of DexKnot, a real-world framework for generalizable bag knotting task with a few demonstrations. Specifically,

\begin{itemize}
\item We propose an imitation learning framework leveraging keypoint representation to enable cross-instance and cross-deformation generalization.

\item We develop a pipeline for keypoint correspondence data collection, using point tracking to avoid massive annotation and physical simulation.

\item We conduct systematic experiments to demonstrate that DexKnot significantly outperforms existing strong baselines on the generalizable bag knotting task.

\end{itemize}

\section{Related Work}
\label{sec:related}

\subsection{Deformable Object Manipulation}

A traditional line of work in deformable object manipulation is model-based methods \cite{li2018learning, yan2021learning, shi2023robocook, wang2023dynamic, li2024deformnet, shi2024robocraft, bauer2024doughnet, zhang2024adaptigraph}, which either build or learn a dynamics model of the object to manipulate. 
The dynamics models can predict the motion and deformation of objects subject to manipulation inputs, facilitating model predictive control (MPC) or model-based reinforcement learning (MBRL). 
Recently, significant advances have been made in end-to-end policy learning.
Model-free reinforcement learning (MFRL) has demonstrated effectiveness in manipulating rope and cloth \cite{matas2018sim, wu2019learning, lin2021softgym}. 
Imitation learning, especially diffusion policy  (DP) \cite{chi2023diffusion}, has also been applied in many relevant tasks, such as garments \cite{wang2025dexgarmentlab}.
Compared to reinforcement learning (RL), DP is easier to train and more friendly for real-world data collection, making it a natural choice for our policy.

Recent advances in physical simulation have largely facilitated deformable object manipulation tasks, including ropes, cloth \cite{matas2018sim, wu2019learning, lin2021softgym, laezza2021reform}, garments \cite{wang2025dexgarmentlab}, tissues \cite{liang2024real}, and plasticine \cite{huang2021plasticinelab, li2023dexdeform}.
Physical simulation is especially crucial for RL, which is expected to learn policies outperforming humans by scalable exploration in virtual environments. However, the sim-to-real gap remains a significant challenge, which is pronounced when objects are highly deformable.

\subsection{Bag Manipulation}
Compared to simple deformable objects, bags present more challenges for manipulation \cite{chen2022autobag, chen2023bagging, bahety2023bag}.
In terms of policy learning, the primary challenge of bag manipulation is generalization for initial deformations.
The state-of-the-art policy learning methods like RL and DP \cite{chi2023diffusion} struggle to generalize with high-dimensional inputs but little data.
To address this, there are some simple yet effective solutions, such as using airflow \cite{xu2022dextairity} or shaking a bag \cite{gu2024shakingbot}.
Another solution is iterative policy, 
which learns to adjust actions iteratively based on visual feedback to achieve precise goal conditions \cite{gao2023iterative, chi2024iterative}.
Our approach leverages representation learning and diffusion policy, enabling generalization by extracting sparse manipulation-relevant keypoints as representation.

In terms of tasks, many of the works in bag manipulation focus on opening a bag or inserting objects into a bag \cite{chen2022autobag}, while less attention is paid to knotting a bag. However, knotting a bag is valuable with applications in many scenarios such as supermarkets. Our framework achieves the knotting task, while not involving any designs specific to knotting. This means our approach is general and can potentially adapt to other tasks.

\begin{figure*}[htbp]
\centerline{\includegraphics[scale=0.5]{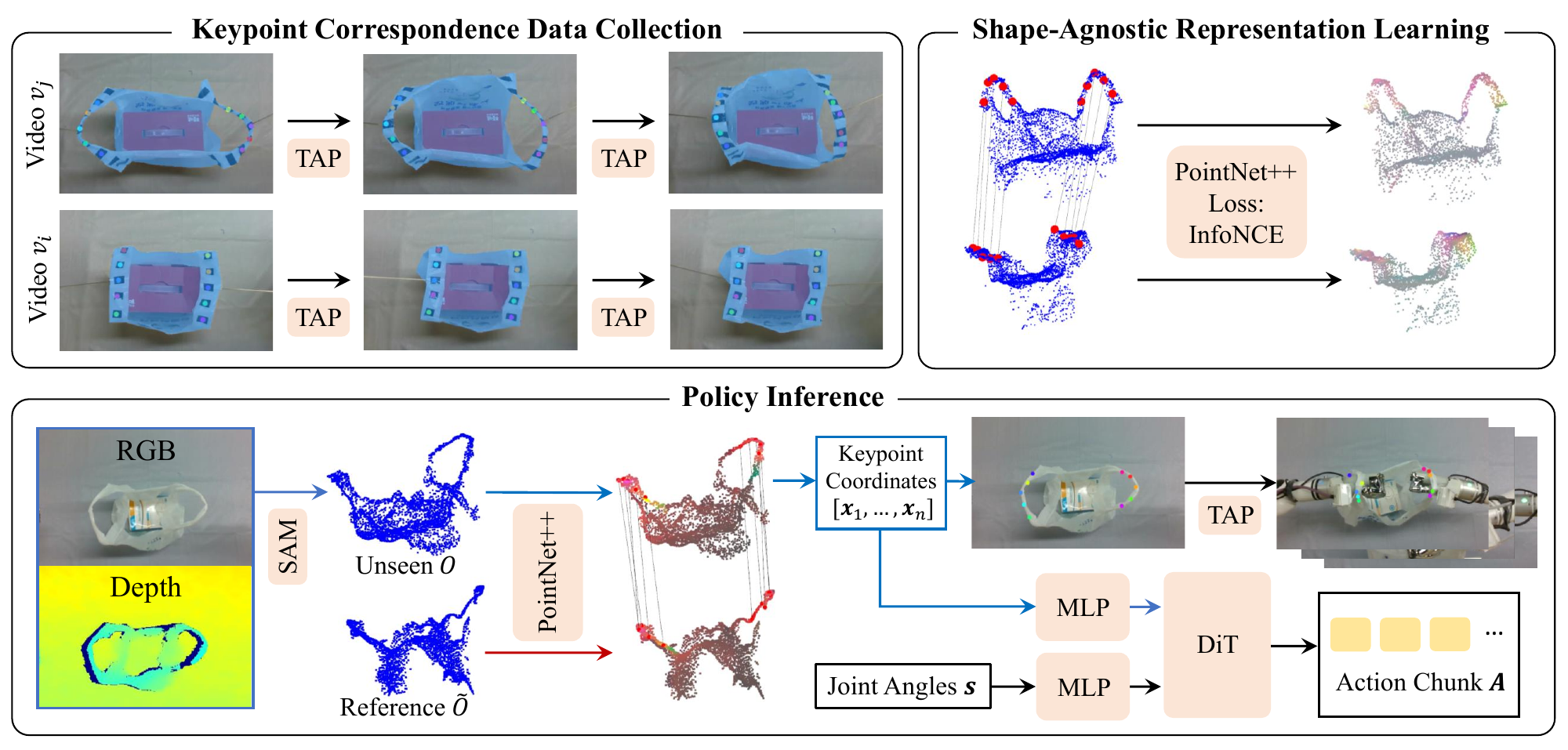}}
\caption{\textbf{Our Proposed Framework.} \textbf{Top left}: For each bag, we perform manual deformation while recording RGB-D videos, and then we track the keypoints for correspondence data construction.
\textbf{Top right}: The PointNet++ encoder learns to produce similar representations for corresponding keypoints across different deformations using an InfoNCE loss. 
\textbf{Bottom row}: During policy inference, keypoints are identified in the initial frame through representation matching and tracked across subsequent frames using TAP. These keypoint coordinates are combined with robot joint states and fed into a Diffusion Transformer to generate an action chunk.
}
\label{fig2}
\end{figure*}

\subsection{Generalizable Visual Representations}

Visual representation aims to encode invariant information across varying situations to facilitate downstream policy.
The most straightforward approach is to simply feed RGB-D image into a U-Net, as employed in diffusion policy \cite{chi2023diffusion}. 
However, such dense representations often contain substantial irrelevant information that can distract the policy and impede generalization.
Point clouds offer a sparser alternative that better captures spatial structure, and recent work 3D diffusion policy \cite{Ze2024DP3} has demonstrated that using point clouds as visual inputs can achieve strong performance and generalization.

Deformable objects exhibit an effectively infinite number of possible states, making it particularly challenging for dense representations to generalize effectively.
In contrast, sparse keypoint representations \cite{grannen2020untangling, wu2024unigarmentmanip} can provide actionable affordance for downstream motion planning or policy learning, facilitating generalization by reducing the dimensionality of observation space.
In this work, we adopt correspondence matching as a powerful method for keypoint identification in novel bag configurations, following its proven effectiveness in garment manipulation \cite{wu2024unigarmentmanip}.

\section{Method}
\label{sec:method}

\subsection{Overview}

Our framework addresses the challenge of generalizable bag knotting by combining representation learning with imitation learning. Despite the infinite degrees of freedom inherent to deformable objects, we leverage the topological consistency of plastic bags to learn shape-agnostic representations. This approach enables identification of sparse keypoints through representation matching for novel bag configurations, significantly reducing observation space dimensionality and thus improving policy generalization.

As shown in \autoref{fig2}, our framework operates through three stages:
\begin{itemize}
\item \textbf{Correspondence Data Collection} (Top left): We perform manual deformation while recording RGB-D videos to capture diverse bag configurations. The keypoints are annotated and tracked to construct correspondence dataset. 

\item \textbf{Shape-Agnostic Representation Learning} (Top right): A PointNet++ encoder learns to produce similar representations for corresponding keypoints across different configurations using contrastive learning with InfoNCE loss.

\item \textbf{Keypoint-Guided Generalizable Policy} (Bottom row): During inference, keypoints are identified through representation matching in the initial frame and tracked across subsequent frames. These coordinates are combined with robot joint states and fed into separate MLPs followed by a Diffusion Transformer (DiT) to generate action chunks for manipulation.
\end{itemize}

This integrated approach enables effective generalization across diverse bag instances and deformation states by leveraging topological consistency while minimizing the observation space through sparse keypoint representation.

\subsection{Correspondence Data Collection}

We develop a pipeline for collecting keypoint correspondence data through real-world manual deformation that avoids both the sim-to-real gap of physical simulation and the burden of extensive manual annotation. 
Each bag is marked with $n$ points, representing keypoints ${p_{key}^{(1)}, ..., p_{key}^{(n)}}$.
Specifically, the $n=10$ keypoints are selected as uniformly distributed points along the handle regions, chosen to capture the essential topological structure relevant to manipulation.
For each configuration (bag instance and initial deformation), we manually deform the bag while recording an RGB-D video $v_i$ using our robot's head-mounted camera, where $i$ denotes the serial number of the video. 
To extract the pixel coordinates of the keypoints from $v_i$, we manually annotate keypoints only in the first frame of each video, then employ Track Any Point (TAP) \cite{doersch2023tapir} to propagate these annotations through subsequent frames.
To segment the bag from the background, we use Segment Anything (SAM) \cite{kirillov2023segment} on the first frame and employ Cutie \cite{cheng2024putting} for mask tracking across frames. 
The resulting data provides rich 3D information: we obtain 3D keypoint coordinates $\boldsymbol{x}=[\boldsymbol{x}_1, ..., \boldsymbol{x}_{n}]$ by combining pixel coordinates with depth information, along with complete point cloud observations $\mathcal{O}$ containing $n_{pc}$ points. 
Finally, we construct our correspondence dataset by randomly matching keypoints across all frames and videos with probability $p_m$, creating positive pairs for contrastive learning while reducing computational burden.
Key hyperparameters are listed in \autoref{tab1}.

\begin{table}[htbp]
\caption{Hyperparameters in the encoder and policy}
\begin{center}
\begin{tabular}{ccc}
\toprule
\textbf{Parameter} & \textbf{Value} & \textbf{Description} \\
\midrule
$n$ & 10 & Number of Keypoints  \\
 $n_{pc}$ & 4096 & Number of Points in Point Cloud  \\
 $p_m$ & 0.001 & Probability of Matching  \\
$d$ & 512 & Encoder Feature Dimension  \\
$D$ & 256 & DiT Input Dimension  \\
$m$ & 150 & Number of Negative Point Samples  \\
 $H$ & 16 & Action Chunk Horizon  \\
\bottomrule
\end{tabular}
\label{tab1}
\end{center}
\end{table}

\subsection{Shape-Agnostic Representation Learning}

We formulate the problem of learning deformation-invariant representations as a contrastive learning task that enforces consistency between corresponding keypoints across different bag configurations. This approach enables our system to recognize the same structural features regardless of how the bag is deformed or which specific instance is being manipulated.
The core objective is to train a feature extractor $F$ that produces identical representations for equivalent keypoints across different point cloud observations. 
Formally, given two point cloud observations $\mathcal{O}^{(1)}$ and $\mathcal{O}^{(2)}$ with corresponding keypoints $p_{key, i}^{(1)}$ and $p_{key, i}^{(2)}$, the representations $F({p_{key, i}^{(1)}})$ and $F({p_{key, i}^{(2)}}) \in \mathbb{R}^{d}$ extracted by the backbone network $F$ should be identical. 
Here we implement $F$ as a PointNet++ \cite{qi2017pointnet++} network, which captures hierarchical spatial features from point clouds while maintaining permutation invariance. 
We normalize all extracted representations to unit vectors,  enabling similarity measurement via dot product: $F({p_{key, i}^{(1)}}) \cdot F({p_{key, i}^{(2)}})$. 
The learning framework follows a contrastive paradigm: for each anchor keypoint $p_{key, i}^{(1)}$ from $\mathcal{O}^{(1)}$, we consider the corresponding point $p_{key, i}^{(2)}$ from $\mathcal{O}^{(2)}$ as the positive sample, while randomly selecting $m$ points ${p^{(2)}_1, p^{(2)}_2, \ldots, p^{(2)}_m}$ from different locations in $\mathcal{O}^{(2)}$ as negative samples. This construction teaches the network to distinguish between equivalent keypoints from other points.
We use InfoNCE \cite{oord2018representation} as the loss function, which has proven effective in contrastive learning scenarios:
\begin{equation}
\label{eq:l_cd}
\mathcal{L} = -\log\left(\frac{\exp(F({p_{key, i}^{(1)}}) \cdot F({p_{key, i}^{(2)}})/\tau)}{\sum_{j=1}^{m} \exp(F({p_{key, i}^{(1)}}) \cdot F({p^{(2)}_j)}/\tau)}\right),
\end{equation}
The temperature $\tau$ modulates the sharpness of the similarity distribution, allowing control over how strongly the model distinguishes between similar and dissimilar pairs.

Given a novel bag configuration, we use SAM to obtain point cloud observation $\mathcal{O}$. To identify keypoints on this novel bag, we employ correspondence matching using a fixed reference observation ${\mathcal{O}^{(ref)}}$—a pre-recorded point cloud of a canonical bag configuration with manually annotated keypoints. This reference remains constant across all inference runs.
For keypoint $p_{key,i}$, we compare the feature representations of all points $\{F(p_{j})\}$ in the novel observation $\mathcal{O}$ against the reference feature $F({p_{key,i}^{(ref)}})$ from  ${\mathcal{O}^{(ref)}}$. The point in $\mathcal{O}$ with the highest similarity is selected as the identified keypoint:
\begin{equation}
p_{key,i}=\mathrm{argmax}(F(p_{j}) \cdot F({p_{key,i}^{(ref)}}))
\end{equation}
\autoref{fig1} and \autoref{fig2} visualize our learned shape-agnostic representation, where corresponding keypoints maintain consistent features across bag instances and deformations. This invariance allows for reliable keypoint identification via correspondence matching on previously unseen bag configurations.

\subsection{Keypoint-Guided Generalizable Policy}
\label{subsec:policy}

Our policy is designed to leverage the keypoint for generalizable bag manipulation. The key idea is that by reducing the observation space to a compact set of geometrically meaningful keypoints, we can learn effective policies that generalize across diverse bag configurations from only a few demonstration data.
The policy operates on the identified keypoint coordinates $\boldsymbol{x}$ obtained through correspondence matching.    
To maintain temporal consistency and avoid reprocessing the entire point cloud at each step, we employ TAP for continuous keypoint tracking, producing updated coordinates $\boldsymbol{x}_t$ at each time step $t$. 
Combined with robot joint angle state $\boldsymbol{s}_t$ as input, the problem can be formulated as learning a policy $\pi$ that effectively models the action distribution $\pi(\cdot | \boldsymbol{s}_t, \boldsymbol{x}_t)$.

We adopt an action-chunking approach with horizon $H$ to improve temporal coherence and enable long-horizon reasoning.
We map keypoint coordinates $\boldsymbol{x}_t$ and robot joint angle $\boldsymbol{s}_t$ into a common embedding space using separate MLPs, yielding ${\boldsymbol{z}}^x_t$ and ${\boldsymbol{z}}^s_t$. 
These embeddings are then stacked to form full observation ${\boldsymbol{z}}^{\text{obs}}_t \in \mathbb{R}^{2 \times D}$.
For action generation, we use a Diffusion Transformer \cite{peebles2023scalable}  to generate multi-step actions following diffusion policy paradigm \cite{chi2023diffusion, liu2024rdt, zhong2025dexgraspvla}. At each time step $t$, we bundle next $H$ actions into a chunk $\boldsymbol{A}_t = \boldsymbol{a}_{t:t+H} = [\boldsymbol{a}_t, \boldsymbol{a}_{t+1}, \ldots, \boldsymbol{a}_{t+H-1}]$. During training, we sample random diffusion step $t_d = k$, and then add Gaussian noise $\boldsymbol{\epsilon}$ to $\boldsymbol{A}_t$ to get the noised action tokens $\tilde{\boldsymbol{A}}_k = \alpha_k \boldsymbol{A}_t + \sigma_k \boldsymbol{\epsilon}$, where $\alpha_k$ and $\sigma_k$ are the standard DDPM coefficients. Next, we feed $\tilde{\boldsymbol{A}}_k$ into DiT with observation feature $\boldsymbol{z}^{\text{obs}}_t$. Each DiT layer performs bidirectional self-attention over action tokens, cross-attention to $\boldsymbol{z}^{\text{obs}}_t$, and MLP transformations, predicting original noise $\boldsymbol{\epsilon}$. By minimizing the discrepancy between the predicted and true noise, the model learns to reconstruct the ground-truth action chunk $\boldsymbol{A}_t$. At inference time, iterative denoising steps recover the intended multi-step action sequence from the learned distribution.

\section{Experiments}
\label{sec:experiments}

\subsection{Experiment Setup}
\label{subsec:setup}

\begin{figure}[htbp]
\centerline{\includegraphics[scale=0.43]{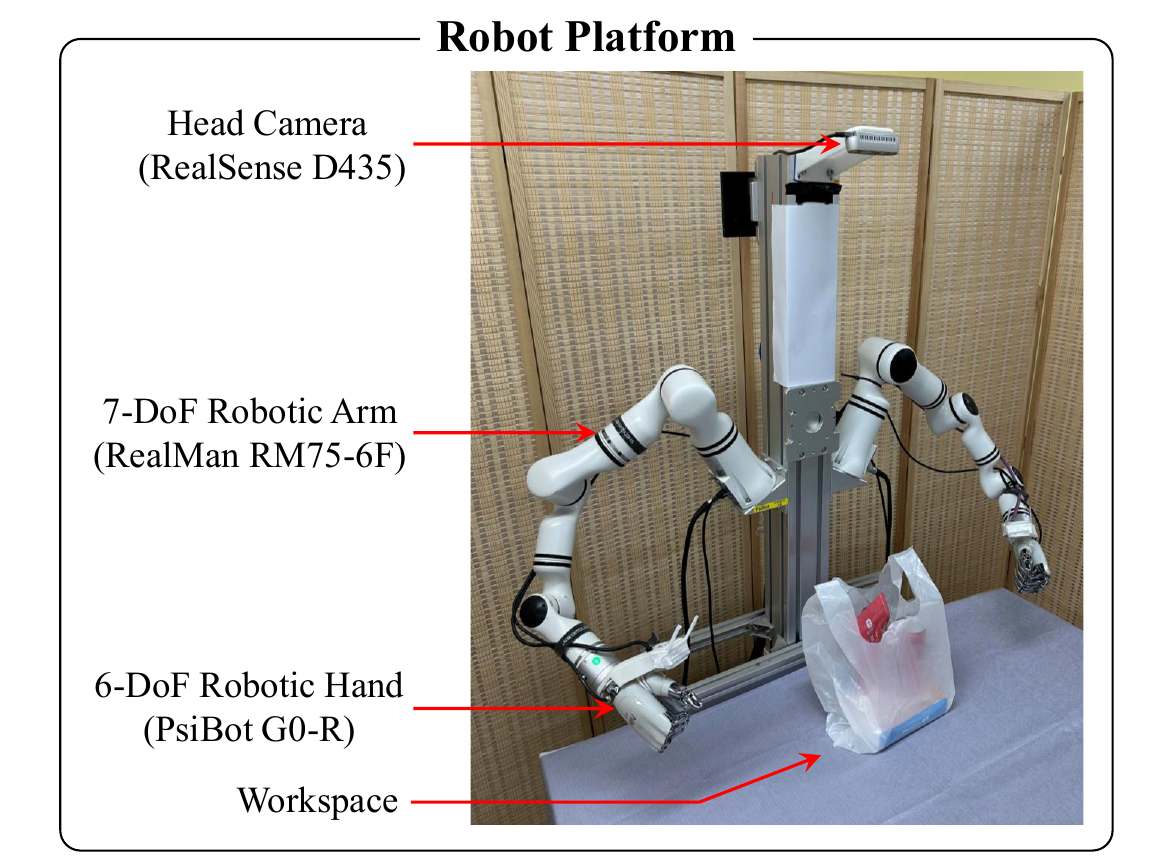}}
\caption{\textbf{Robot setup}. Our robot platform includes a RealMan RM75-6F dual-arm system with PsiBot G0-R 6-DoF dexterous hands and a head-mounted Intel RealSense D435 RGB-D camera. }
\label{fig3}
\end{figure}

\textbf{Robot Platform.} All data collection and evaluation experiments are conducted using RealMan RM75-6F dual-arm robot equipped with PsiBot G0-R dexterous hands and a head-mounted Intel RealSense D435 RGB-D camera (\autoref{fig3}). 
Although a wrist camera is available on the platform, it is not used in our current framework, 
as the head camera provides sufficient visual coverage for keypoint identification and tracking. Actions are recorded and executed at approximately 10 Hz. 
The action space consists of the joint angles of both arms and dexterous hands, yielding a total of 26 DoF: 7 DoF per arm and 6 DoF per hand.

\begin{figure*}[htbp]
\centerline{\includegraphics[scale=0.495]{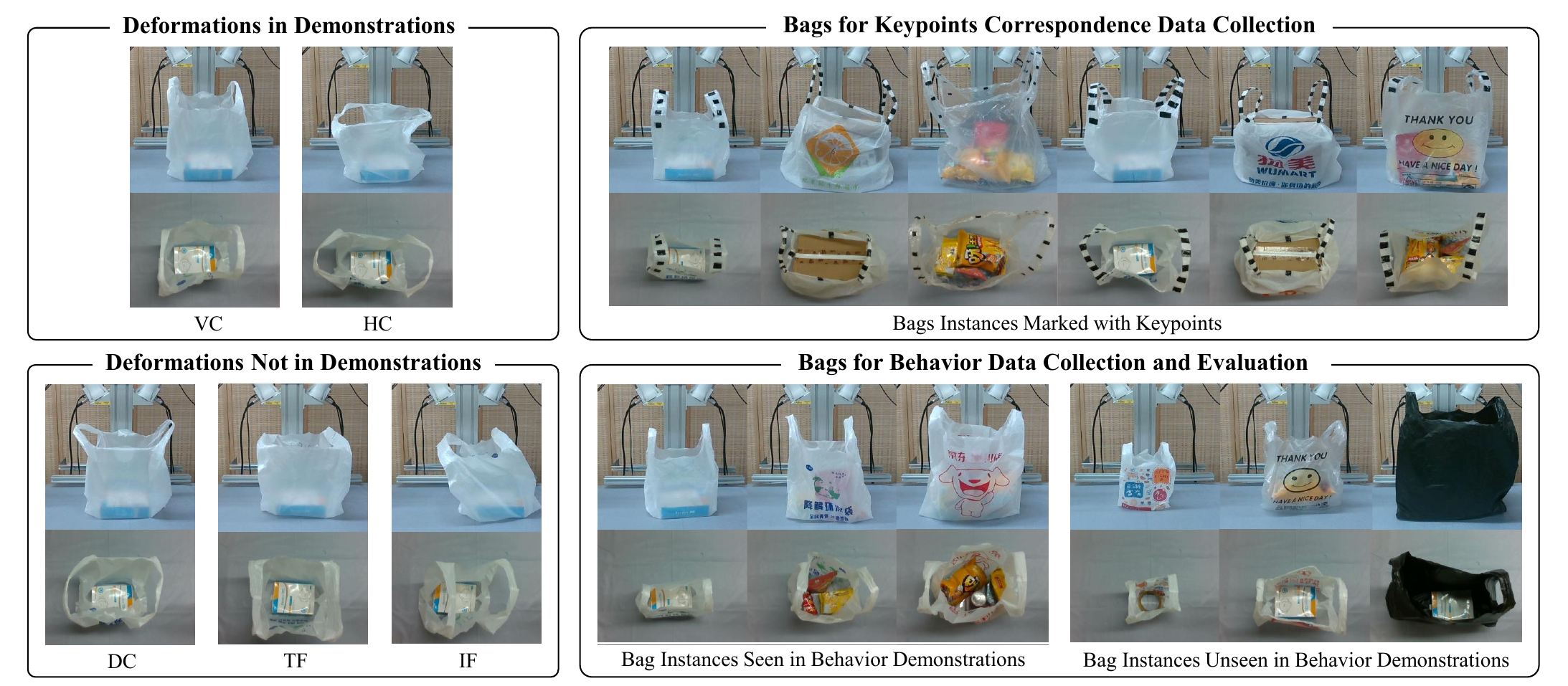}}
\caption{\textbf{Bag deformations and instances}. 
\textbf{Top left}: Deformations included in behavior demonstrations. \textbf{Bottom left}: Deformations not included in behavior demonstrations.
\textbf{Top right}: bags used for keypoint correspondence data collection. \textbf{Bottom right}: bags used for behavior demonstration data collection and novel bags for cross-instance evaluation that are not included in the keypoint correspondence data or behavior demonstrations.
}
\label{fig4}
\end{figure*}

\textbf{Deformation State Definitions.} 
We define five distinct deformation states to standardize data collection and evaluation (\autoref{fig4}, left column):

\begin{itemize}
\item Vertical-Compressed (VC): The handles are oriented vertically and in a compressed rope-like state.
\item Horizontal-Compressed (HC): The handles are oriented horizontally and in a compressed rope-like state.
\item Diagonal-Compressed (DC): The handles are oriented diagonally and compressed into a rope-like state, which can be considered as an interpolated state of VC and HC.
\item Twisted-Flat (TF): The handles are twisted inward and splayed flat.
\item Inclined-Flat (IF): The handles lean to one side and splayed flat.
\end{itemize}
These deformation states are consistently used across all data collection and evaluation procedures.

\textbf{Data.} 
For keypoint correspondence data, we use six plastic bags of varying sizes and shapes, each marked with $n=10$ keypoints on handles (\autoref{fig4}, top right). 
Note that there are some additional markers on the opening of some bags, which are not used as keypoints. 
Two experimenters manually deform each bag while the head-mounted camera records the manual deformation process. 
In total, we collect 117 videos across all six bags. From these videos, any two frames are randomly matched with probability $p_m$, yielding approximately 15000 frame pairs for contrastive learning.

For behavior demonstrations, we use three bags: two are new bags but of the types present in the correspondence data and one novel type (\autoref{fig4}, bottom right, Bag Instances Seen in Behavior Demonstrations).  
All bags used in this stage have no markers. A knotting action involves four stages: threading handles; hook the left inner handle with the right index finger and thumb; hook the right outer handle with the left index finger and thumb; tightening the knot.
We collected 54 human demonstration trajectories, each comprising 160 action steps, across two initial deformation states (\autoref{fig4}, top left):

\textbf{Evaluation Protocol.} We evaluate generalization across initial deformations and bag instances.
For cross-deformation generalization, each bag is evaluated in five states: VC and HC, which are present in the demonstrations; DC, TF, and IF, which are not present in the demonstrations (\autoref{fig4}, bottom left).
For cross-instance generalization, policies are tested on three bags present in demonstrations (\autoref{fig4}, bottom right, Bag Instances Seen in Behavior Demonstrations) and three novel bags not present in demonstrations (\autoref{fig4}, bottom right, Bag Instances Unseen in Behavior Demonstrations).

\textbf{Metric.} 
We report success rates as the number of successful trials divided by the total attempts across all test conditions.

\textbf{Baselines.} We compare against state-of-the-art imitation learning approaches and Vision-Language-Action model:
\begin{itemize}
\item \textbf{DP:} Standard Diffusion Policy \cite{chi2023diffusion} trained on our demonstration data with raw RGB images as input.
\item \textbf{DP3:} 3D Diffusion Policy trained on our demonstration data with bag point clouds as input.
\item\textbf{$\pi_{0}$:} Vision-Language-Action model $\pi_{0}$ \cite{black2024pi_0} fine-tuned on our demonstration data. 
\end{itemize}

\subsection{Generalization Evaluation}

We evaluate the generalization capability of our approach against baseline methods across five initial deformations for bag instances seen and unseen by the demonstrations.
We note that $\pi_{0}$ performs poorly even on seen bags and often results in hand collisions, so we omit further tests on unseen bags. This is likely attributed to the embodiment mismatch: $\pi_{0}$ is pre-trained on data collected with wrist and base cameras, whereas our setup uses only a single head-mounted RGB-D camera. Combined with limited fine-tuning data, this mismatch likely contributes to $\pi_{0}$'s degraded performance in our setting.
The standard DP approach demonstrates limited performance as well, likely due to the high dimensionality of raw RGB input and the absence of depth information. Consequently, our primary comparative analysis focuses on DP3, which serves as our main baseline due to its strong performance.

\autoref{tab2} shows the success rates of DexKnot and baselines on bag instances seen in demonstrations. 
For VC and HC (seen deformations) and DC (interpolation deformation), both DP3 and DexKnot achieve high success rates. 
For TF and IF (out-of-distribution deformations), our approach significantly outperforms DP3, demonstrating better generalization to novel deformations. 
The results can be explained as follows:
The flat handles are never seen by DP3's encoder, thus leading to wrong behavior of the policy; In contrast, the keypoints on the handles can still be identified by DexKnot's encoder since it has been pretrained with diverse states during manual deformation. 
This performance gap is particularly evident for the IF case: The point cloud deviates a lot from the training data, which cannot be handled by DP3's encoder; In contrast, DexKnot can still identify the keypoints, enabling the policy to perform the task.

\begin{table}[htbp]
\caption{Results on Seen Bags across Deformations}
\begin{center}
\begin{tabular}{ccccc}
\toprule
\textbf{Methods} & \textbf{VC \& HC} & \textbf{ DC} & \textbf{TF} & \textbf{IF} \\
\midrule
DP  & 3/18 & 2/9 & 1/9 & 2/9 \\
DP3  & \textbf{17/18} & \textbf{9/9} & 2/9 & 0/9 \\
$\pi_{0}$  & 1/18 & 0/9 & 1/9 & 0/9 \\
Ours  & 16/18 & 8/9 & \textbf{8/9} & \textbf{4/9} \\
\bottomrule
\end{tabular}
\label{tab2}
\end{center}
\end{table}

\autoref{tab3} shows the success rates of DexKnot and baseline methods for bag instances that are never present in the correspondence data or behavior demonstrations.
While all methods exhibit reduced performance compared to seen instances, DexKnot significantly outperforms DP3 across all initial deformations, particularly excelling in twisted and inclined cases. 
These results demonstrate that our approach not only generalizes better to novel deformations but also maintains more consistent performance when presented with unseen instances.

\begin{table}[htbp]
\caption{Results on Unseen Bags across Deformations}
\begin{center}
\begin{tabular}{ccccc}
\toprule
\textbf{Methods} & \textbf{VC \& HC} & \textbf{ DC} & \textbf{TF} & \textbf{IF} \\
\midrule
DP  & 4/18 & 1/9 & 0/9 & 0/9 \\
DP3  & 14/18 & 6/9 & 1/9 & 0/9 \\
Ours  & \textbf{15/18} & \textbf{8/9} & \textbf{6/9} & \textbf{4/9} \\
\bottomrule
\end{tabular}
\label{tab3}
\end{center}
\end{table}

\begin{figure*}[htbp]
\centerline{\includegraphics[scale=0.53]{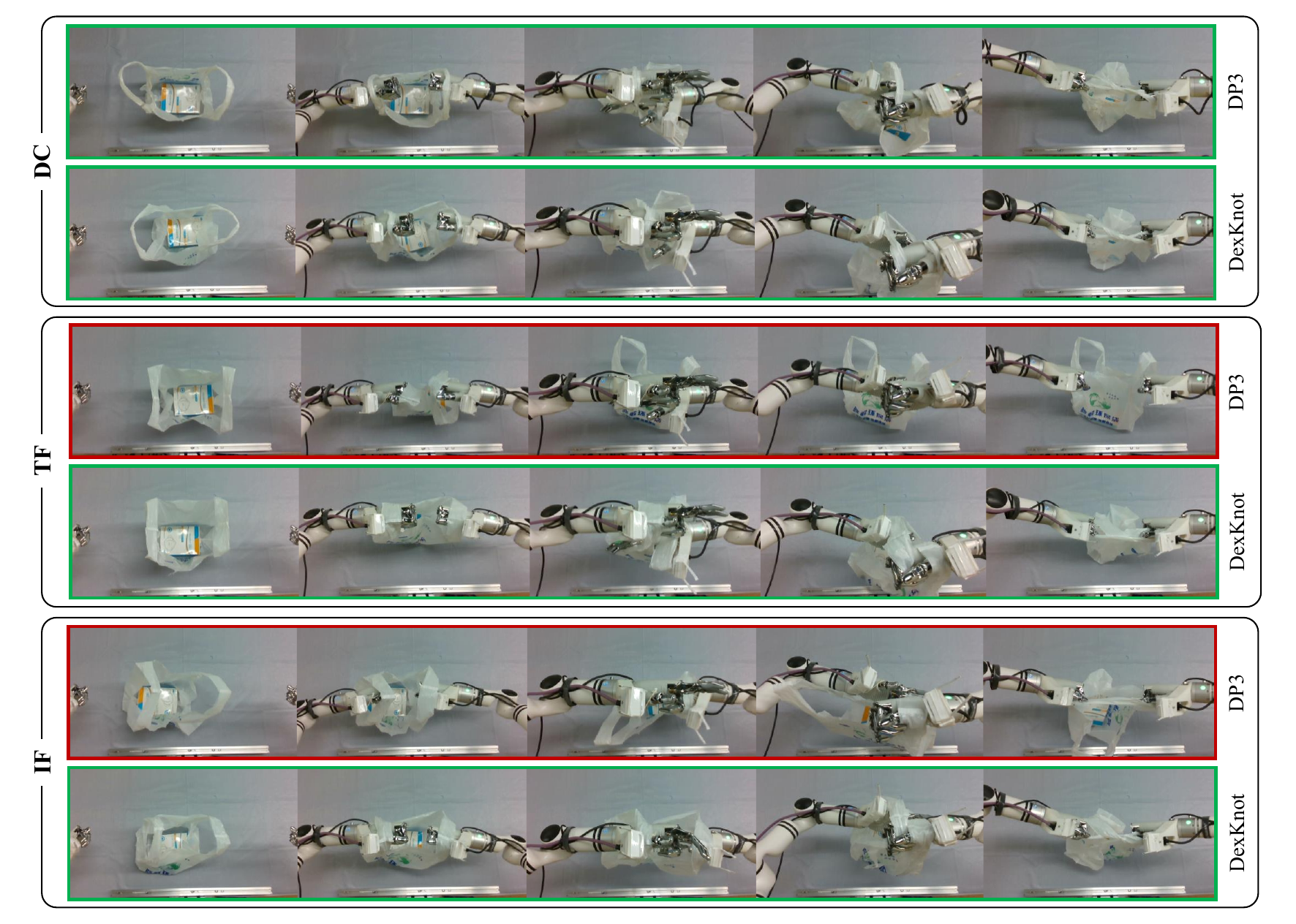}}
\caption{\textbf{Qualitative comparison of policy executions}. Successes and failures are indicated by green and red bounding boxes, respectively. 
\textbf{Top row}: Both DP3 and DexKnot successfully complete the knotting task under Diagonal-Compressed (DC) deformation conditions. 
\textbf{Middle row}: In Twisted-Flat (TF) conditions, DP3 fails to thread the handle while DexKnot successfully accomplishes the task.
\textbf{Bottom row}: In Inclined-Flat (IF) conditions, DP3 fails to thread the handle while DexKnot successfully accomplishes the task.
}
\label{fig5}
\end{figure*}

\autoref{fig5} shows qualitative comparisons between DP3 and our approach for a seen bag in three initial deformations. While both methods successfully completed the knotting task under DC configuration, DP3 failed to identify handle locations in TF and IF configurations, leading to the failure of the task. In contrast, DexKnot maintained robust performance across all deformations.

Our results indicate that while DP3 and DexKnot show comparable performance on seen bag instances, seen deformations, and simple interpolated deformation, our approach demonstrates significantly better generalization when presented with novel deformations, such as inclined and twisted states.

\subsection{Ablation Studies}
\label{subsec:ablation}

To evaluate the contribution of key components in DexKnot, we conducted ablation studies comparing our full framework against two ablated versions on bag instances unseen in demonstrations:

\begin{itemize}
\item \textbf{Ours w/o TF/IF:} This variant removes exposure to twisted and inclined deformations during the encoder's training phase, testing the importance of diverse manual deformations for learning shape-agnostic representations.
\item \textbf{Ours w/o TAP:} This variant replaces the TAP-based keypoint tracking with an alternative approach: using Cutie to track the bag's mask and identifying the keypoints by the encoder at each step. 
\end{itemize}

As quantitatively demonstrated in \autoref{tab4}, both ablated versions show performance degradation across all deformations compared to the full method. The performance drop in \textbf{Ours w/o TF/IF} indicates that training the encoder on a diverse set of deformations is crucial for learning shape-agnostic representations that enables generalization in the downstream policy. 
The inferior results of \textbf{Ours w/o TAP} indicates that identifying keypoints initially and then tracking them provides more reliable state estimation than tracking the mask and identifying the keypoints in each frame. 
These results validate the importance of each component in our complete framework.

\begin{table}[htbp]
\caption{Ablation Study Results on Unseen Bags}
\begin{center}
\begin{tabular}{ccccc}
\toprule
\textbf{Methods} & \textbf{VC \& HC} & \textbf{ DC} & \textbf{TF} & \textbf{IF} \\
\midrule
Ours w/o TAP  & 13/18 & 7/9 & 5/9 & \textbf{4/9} \\
Ours w/o TF/IF & \textbf{17/18} & 7/9 & 1/9 & \textbf{4/9} \\
Ours  & 15/18 & \textbf{8/9} & \textbf{6/9} & \textbf{4/9} \\
\bottomrule
\end{tabular}
\label{tab4}
\end{center}
\end{table}

\section{Conclusion}
\label{sec:conclusion}

We present DexKnot, a framework that integrates shape-agnostic representation learning with diffusion policy for generalizable bag knotting. By encoding crucial manipulation information into a sparse set of keypoints, this approach dramatically reduces the observation space dimensionality, enabling robust generalization to both unseen initial deformations and bag instances. 
Experimental results demonstrate superior performance over baseline methods, particularly for novel deformations.
While demonstrated on bag knotting, DexKnot's pipeline could extend to other deformable object manipulation tasks (e.g., fabric manipulation) where objects have consistent topological structure. 
We leave the exploration of such extensions to future work.

Despite the advantages, DexKnot also has some limitations. 
First, although our correspondence data collection pipeline significantly reduces manual effort by requiring only first-frame annotations, the initial annotation requirement remains a notable limitation. 
Second, the keypoint representation's low dimensionality, while beneficial for generalization, introduces a vulnerability to misidentification errors. This represents an inherent trade-off between representations' sparsity and robustness that warrants further investigation.

\bibliographystyle{IEEEtran}  
\bibliography{refs}           

\end{document}